\documentclass{sig-alternate}
\usepackage{natbib}
\usepackage{amsmath}
\usepackage{tikz}
\usepackage{subcaption}
\usepackage{caption}
\usetikzlibrary{fit,positioning}

\begin{document}
%

\title{Clustering via Content-Augmented \\ Stochastic Blockmodels}
%
%
%
%
%

\numberofauthors{5} 
%
\author{
%
%
\alignauthor
J. Massey Cashore\titlenote{This work was done while the author was at Cornell University}\\
       \affaddr{University of Waterloo}\\
       \affaddr{Waterloo, ON, Canada}\\
       \email{jmcashor@uwaterloo.ca}
\alignauthor
Xiaoting Zhao\\
       \affaddr{Cornell University}\\
       \affaddr{Ithaca, NY 14850}\\
       \email{xz337@cornell.edu}
\alignauthor 
Alexander A. Alemi\\
       \affaddr{Cornell University}\\
       \affaddr{Ithaca, NY 14850}\\
       \email{aaa244@cornell.edu}
\and  
\alignauthor Yujia Liu\\
       \affaddr{Cornell University}\\
       \affaddr{Ithaca, NY 14850}\\
       \email{yl2254@cornell.edu}
\alignauthor 
Peter I. Frazier\\
       \affaddr{Cornell University}\\
       \affaddr{Ithaca, NY 14850}\\
       \email{pf98@cornell.edu}
}
\maketitle
\begin{abstract}
Much of the data being created on the web contains interactions between
users and items.  Stochastic blockmodels, and other methods for community detection and clustering of bipartite graphs, can infer latent user communities and latent item clusters from this interaction data.  These methods, however, typically ignore the items' contents and the information they provide about item clusters, despite the tendency of items in the same latent cluster to share commonalities in content.  We introduce content-augmented stochastic blockmodels (CASB), which use item content together with user-item interaction data to enhance the user communities and item clusters learned.  Comparisons to several state-of-the-art benchmark methods, on datasets arising from scientists interacting with scientific articles, show that content-augmented stochastic blockmodels provide highly accurate clusters with respect to metrics representative of the underlying community structure.

\end{abstract}

\category{H.2.8}{Database Management}{Database Applications}[Data Mining]
\category{I.5.3}{Pattern Recognition}{Clustering}[Algorithms]

\terms{Algorithms}

\keywords{Clustering, Community Detection}

\section{Introduction}

Motivated by an application to the arXiv \cite{arxiv},
we consider the problem of finding hard clusters of scientific 
articles in the presence of user-interaction data and document
content. To this end, we develop a generative model of user-item
interaction as well as item content, such that each document
is associated with a single scalar latent variable, indicating 
its cluster membership.
Our model is essentially a stochastic blockmodel applied to the
bipartite user-item interaction graph, combined with a content distribution
on the document vertices.

The application to arXiv motivating this work is that of developing a finer-grained categorization of papers than currently exists, which will be used to offer users more fine-grained control over the daily and weekly feeds of newly submitted papers, and within a new information filtering system that will learn user preferences \cite{ZhaoFrazier2014}. The primary user base will be those who frequently visit the website to stay up-to-date with the research community, as is common among physics researchers.

Stochastic blockmodels were introduced to discover latent community structure in graphs \cite{holland1983stochastic},
typically formed by people or other entities interacting with each other (each person is a node, and edges indicate interactions),
or by people interacting with text documents, images, videos, or some other object 
(each person is a node, each object is a node, and edges indicate interactions, forming a bipartite graph).
In this second kind of application, interaction information is the only information typically used, and information from the documents themselves is ignored.

Traditional bipartite stochastic blockmodels assume that different communities tend to interact differently with each text document, with some communities tending to interact more frequently with a given document type, and other communities interacting more frequently with other document types.  
This differential preference of communities for documents induces a latent document clustering, with bipartite stochastic blockmodels attempting to learn this latent clustering from interaction information alone. Our model adds an additional assumption: that documents in each cluster have distinct characteristics, observable in the words that occur in them.  When this assumption is satisfied, we argue that it can and should be used to improve performance.

While this assumption does not necessarily hold in all community detection applications, we argue that it holds in a wide variety of settings.
In this paper, we apply this model to scientists interacting with scientific articles, 
where the words that tend to appear in articles preferred by a community vary considerably from community to community.
Our model could also be applied to communities interacting with other kinds of items, e.g., videos, but in our empirical studies we focus on text.

Our model can also be seen as a co-clustering algorithm, because it provides a clustering of both users and documents.
However, our model is distinguished from all other co-clustering algorithms of which we are aware, in that it uses
not just the interaction information, but also co-variates observable in the documents.
Thus, our model is distinguished from co-clustering approaches that use only document comments (e.g., based only the matrix of word co-occurrence) by the way it takes advantage of user co-access to find the mapping of contents to clusters that matches the communities' preferences.
It is distinguished from co-clustering approaches that only use user interactions in that document contents are used to refine and improve the co-clustering.

An additional advantage of including document covariates into our model is new documents with no interaction history can be included into an appropriate document cluster, addressing the cold-start problem.

There has been growing interest in combining user interaction and item content
in the context of recommender systems \cite{WangBlei2011, Claypool1999,Balabanovic1997,
Salter2006,Basilico2004,melville2002content}, as well as combining citations and content
in the context of community discovery (specifically on document networks) \cite{NallapatiMMSBLDA2008,
CombineLinkContent2009, GraphClusteringAttributes2009}. To our knowledge, 
this paper is the first to approach the problem of clustering as a community detection task
on the network of user-document interactions.

In section 2 we provide a detailed description of the model, and in
section 3 we describe how variational methods can be used for
inference. In section 4 we apply the model to two real-world datasets,
from the arXiv, and compare them to a several baseline clusters.

\section{Content-Augmented Stochastic Blockmodels}
Suppose we are dealing with an application on the web such that there
are $D$ items and $U$ users potentially interested in the items.
Suppose that over time the users have been shown and provided feedback
on a subset of items. We encode their feedback with the variable $Y$,
defined by
\[
Y_{ij}=\left\{ \begin{array}{cc}
1 & \mbox{if item \ensuremath{i} was shown to user \ensuremath{j} and relevant}\\
0 & \mbox{if item \ensuremath{i} was shown to user \ensuremath{j} and irrelevant}\\
\triangle & \mbox{if item \ensuremath{i} was not shown to user \ensuremath{j}.}
\end{array}\right.
\]
Note we only consider a binary response (relevant / irrelevant), and
use the symbol $\triangle=Y_{ij}$ to denote the case where item $i$
is not shown to user $j$.

The model proceeds by assuming there are $k_{d}$ clusters such that
each item $i$ belongs to cluster $z_{i}\in\{1,\dots,k_{d}\}$ and
there are $k_u$ communities such that each user $j$ belongs to
community $w_{j}\in\{1,\dots,k_u\}$. We assume the community membership
of a user and cluster membership of a paper completely determines
the probability the user finds the paper relevant. Explicitly, the
model assumes that
\begin{equation}
p(Y_{ij}=1|z_{i}=x,w_{j}=y,Y_{ij}\ne\triangle)=q_{xy}
\end{equation}
for constants $q_{xy}$ ranging over item clusters $x$ and user communities
$y$. For simplicity, we encode the $q_{xy}$ as a matrix $Q=[q_{xy}]$. We assume
the observed $Y_{ij}$ are all sampled independently.

Finally, we endow the latent variables described above  with
the following Bayesian priors:
\begin{itemize}
\item The cluster $z_{i}$ of item $i$ follows a uniform distribution on
$k_{d}$ elements.
\item The community $w_{j}$ of user $j$ follows a uniform distribution
on $k_{u}$ elements.
\item The cluster-community interest probabilities $q_{xy}$ follow a $\mathrm{Beta}(\alpha,\beta)$
for some $\alpha,\beta>0$.
\end{itemize}
We further simplify things by assuming that $k_u=k_d$, that is the number of user
communities and document clusters is equal, and use the variable
$K$ to indicate this value.

The model described up to this point is a stochastic blockmodel on a bipartite
graph, without any notion of item content.

To augment the model with content,
we suppose that each item can be represented by an $F$-dimensional
feature vector, such that the $n$th entry counts how many time the
$n$th trait occurred, for some set of $F$ traits. 
Let $d_{i}$ represent the feature vector for the $i$th item.

In order to force that items in the same cluster should be
similar in content, we assume that associated with each item cluster
$x$ is a probability vector $p_{x}\in[0,1]^{F}$, $\sum_{\ell}p_{x\ell}=1$,
such that if item $i$ is in cluster $x$ then the feature vector
$d_{i}$ is created from $N_{i}$ samples of a $\mathrm{Multinomial}(p_{x})$
distribution (the process by which $N_{i}$ is chosen is unimportant). 
We assume the observed $d_{i}$ are all sampled
independently, and we place a $\mathrm{Dirichlet}(\gamma)$ prior on each $p_x$, for some
$\gamma\in\left(\mathbb{R}_{>0}\right)^{F}$.

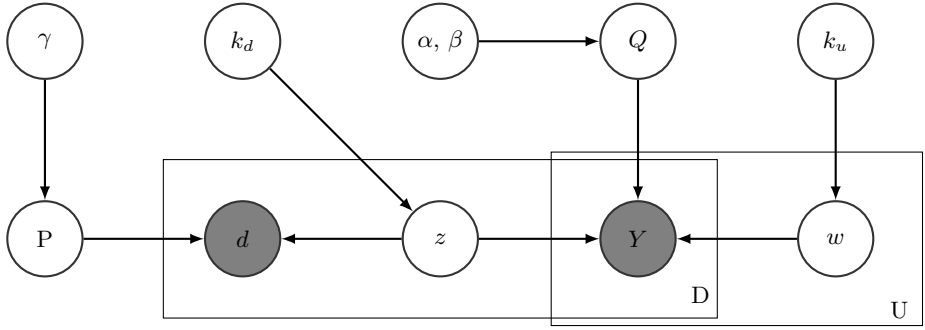
\begin{figure*}
\centering
\begin{tikzpicture}
\tikzstyle{main}=[circle, minimum size = 10mm, thick, draw =black!80, node distance = 16mm]
\tikzstyle{connect}=[-latex, thick]
\tikzstyle{box}=[rectangle, draw=black!100]
  \node[main] (gamma) [] {$\gamma$ };
  \node[main](p)[below=of gamma] {P};
  \node[main, fill=black!50](d)[right=of p]{$d$};
 \node[main](z)[right=of d]{$z$};
\node[main, fill=black!50] (Y) [right=of z] {$Y$};
  \node[main] (w) [right=of Y] { $w$};
  \node[main](priorZ)[above=of d] {$ k_d$};
  \node[main] (q) [above=of Y] {$Q$ };
 \node[main](beta)[above=of z]{$\alpha$, $\beta$};
  \node[main](priorW)[above=of w] {$k_u$};
  \path (gamma) edge [connect] (p)
        (p) edge [connect] (d)
        (priorZ) edge [connect] (z)
        (z) edge [connect] (d)
	(z) edge [connect] (Y)
	(priorW) edge [connect](w)
	(w) edge[connect](Y)
   	(q) edge [connect] (Y)
   	(beta) edge [connect] (q);
  \node[rectangle, inner sep=0mm, fit= (d) (z) (Y),label=below right:D, xshift=27mm] {};
  \node[rectangle, inner sep=5.4mm,draw=black!100, fit= (d) (z) (Y)] {};
  \node[rectangle, inner sep=2mm, fit= (Y) (w),label=below right:U, xshift=12mm] {};
  \node[rectangle, inner sep=6.4mm, draw=black!100, fit = (Y) (w)] {};
\end{tikzpicture}
\caption{Graphical representation of the content-augmented model}
\end{figure*}

\begin{table*} 
\centering
\caption{Expected Natural Parameters for Complete Conditionals and Relevant Expectations. Entries in the ``Relevant Expectations'' column can be used to compute entries in the ``Expected Natural Parameters'' column. $\Psi$ represents the digamma function.}
\begin{tabular}{|c|c|c|c|} \hline
Var & Parameter & Expected Natural Parameter & Relevant Expectation\\ \hline
$z_i$ & 
$\phi_{i, x}, x \in \{1, ..., k_d\}$ &
$\mathbb{E}_q\left[\ln \left[ \prod_{\ell=1}^{F}p_{z_{i}\ell}^{d_{i\ell}} \prod_{j:Y_{ij}=1}q_{z_{i}w_{j}} \prod_{j:Y_{ij}=0}(1-q_{z_{i}w_{j}}) \right]\right]$ &
$\text{Pr}[z_{i}=x]=\phi_{i,x} $
 \\ \hline
$w_j$ & 
$\varphi_{i, y}, y \in \{1, ..., k_u\}$ &
$\mathbb{E}_q\left[\ln \left[\prod_{i:Y_{ij}=1}q_{z_{i}w_{j}}\prod_{i:Y_{ij}\ne1}(1-q_{z_{i}w_{j}})\right]\right]$&
$\text{Pr}[w_{j}=y]=\varphi_{j,y} $
\\ \hline
$q_{xy}$ &
$\alpha_{xy}, \beta_{xy}$ &
$\alpha-1+\mathbb{E}_q\left[\sum_{(i,j):z_{i}=x,w_{j}=y}Y_{ij}\right]$, &
$\mathbb{E}[\ln q_{xy}] = \Psi(\alpha_{xy})- \Psi(\alpha_{xy}+\beta_{xy})]$ 
\\
&
&
$\beta-1+\mathbb{E}_q \left[\sum_{(i,j):z_{i}=x,w_{j}=y}(1-Y_{ij})\right]$ &
$\mathbb{E}[\ln(1-q_{xy})] = \Psi(\beta_{xy})- \Psi(\alpha_{xy}+\beta_{xy})]$
\\
 \hline
$p_x$ &
$\gamma_{x\ell}, \ell \in \{1,...,F\}$ &
$\gamma_{\ell}-1+\mathbb{E}_q \left[\sum_{i:z_{i}=x}d_{i\ell}\right]$  &
 $\mathbb{E}[\log p_{x\ell}] = \Psi(\lambda_{x\ell})-\Psi(\sum_{l}\lambda_{xl})$ 
\\
\hline\end{tabular}
\label{table:natparams}
\end{table*}

This fully describes the content-augmented stochastic blockmodel (CASB).
In the following section we describe how variational inference and Gibbs sampling
can be used to infer the latent variables.
A graphical depiction of the
CASB can be seen in Figure 1, illustrating the dependencies between all
latent variables.

\subsection{Related Work}

There is growing literature at the intersection of clustering and community-detection,
as well as combining community-detection approaches with node attributes (content).
In the context of document networks, the links considered are most often citations,
as opposed to user interactions.

In \cite{NallapatiMMSBLDA2008} the authors introduce two models which jointly
describe text and citations.  The first combines latent Dirichlet allocation
\cite{Blei2003LDA} with mixed-membership stochastic blockmodels 
\cite{airoldi2009mixed}.  They find
this model leads to relatively intractable inference.  In response, they
introduce another model called Link-PLSA-LDA which 
associates a multinomial distribution with each
article, from which the articles citations are drawn. Both models use the graph
structure and article content to learn latent \emph{vector representations} for each
of articles.  While powerful, these do not immediately lend themselves to hard clusters.

The problem of clustering an arbitrary graph with node attributes is studied
in \cite{GraphClusteringAttributes2009}. Rather than taking a probabalistic approach
the authors augment the underlying graph by adding a node for each attribute, with
links to each of the original nodes containing the attribute.  Vertex closeness
is given by a neighborhood random walk model, and the clusters are computed
via the resulting distance function.

In \cite{CombineLinkContent2009} the problem of community detection is approached
by introducing a discriminiative model combining link and content information.
Initially they introduce the Popularity-based Conditional Link model (PCL).  PCL
assumes there is a latent variable describing each node's community membership,
and the probability of a link between two nodes depends on each node's popularity
and community membership. Content is introduced into the model by setting the probability
of belonging to a specific community as the exponential of a linear function of the
node's content vector.

\section{Inference}

It is intractable to directly optimize the likelihood of the latent variables.
Instead, we appeal to variational inference techniques to find approximate
estimates of the latent variables \cite{wainwright2008variational}. In variational
inference, we associate to each latent variable a family of variational distributions 
each parameterized by a free variational parameter. These parameters are 
then optimized to find the closest member of the family to the posterior (in terms
of KL-divergence). We will use mean-field variational
inference, which assumes that the complete joint variational distribution factors.

We assume the following variational parameters and distributions:
\begin{eqnarray}
z_i | \phi_i \sim \mathrm{Multinomial}(\phi_i),\\
w_j | \varphi_j \sim \mathrm{Multinomial}(\varphi_j),\\
q_{xy}|\alpha_{xy}, \beta_{xy} \sim \mathrm{Beta}(\alpha_{xy}, \beta_{xy}), \\
p_x|\lambda_x \sim \mathrm{Dirichlet}(\lambda_x).
\end{eqnarray}

Let $q$ represent the distribution defined above. For notational convenience we
will write expressions involving $q$ with the understanding that the distribution
is conditioned on the variational parameters.

Recall the evidence lower
bound (ELBO) $\mathcal{L}(q)$ is defined as
\begin{equation}\mathcal{L}(q)=\mathbb{E}_q[\log p(d,Y,z,w,P,Q)] - 
                   \mathbb{E}_q[\log q(z,w,P,Q)] 
\label{eq:elbo}
\end{equation}
and is equal to the KL-divergence between $q$ and the posterior, up
to an additive constant. In order to find the optimal variational parameters
we optimize $\mathcal{L}(q)$ using coordinate ascent. Following  \cite{hoffman2013variational,
wainwright2008variational} the update for each variational parameter in coordinate ascent
equals the variational expectation of the natural parameter of the complete conditional 
corresponding to the relevant latent variable.
The complete conditional distributions for each of the latent variables are:
\begin{align} \label{zcomplete}
\begin{split}
p(z_{i}|z_{-i},w,P,Q,d,Y) \propto & \prod_{\ell=1}^{F}p_{z_{i}\ell}^{d_{i\ell}}\prod_{j:Y_{ij}=1}q_{z_{i}w_{j}} \\
                                  & \prod_{j:Y_{ij}=0}(1-q_{z_{i}w_{j}})
\end{split}
\end{align}
\begin{equation} \label{wcomplete}
p(w_{j}|w_{-j},z,P,Q,d,Y)\propto\prod_{i:Y_{ij}=1}q_{z_{i}w_{j}}\prod_{i:Y_{ij}\ne1}(1-q_{z_{i}w_{j}})
\end{equation}

\begin{flalign} \label{qcomplete}
\begin{split}
q_{xy}|Q_{-xy},w,z,P,d,Y \sim \mathrm{Beta}\left(\alpha', \beta'\right)
\end{split}
\end{flalign}
\begin{equation} \label{pcomplete}
p_{x}|P_{-x},w,z,Q,d,Y\sim\mathrm{Dirichlet}(\gamma')
\end{equation}

The parameters in equations (\ref{qcomplete}) and (\ref{pcomplete}) are given by:
\begin{align}
\alpha' &= \alpha+\sum_{(i,j):z_{i}=x,w_{j}=y}Y_{ij} \\
\beta' &= \beta+\sum_{(i,j):z_{i}=x,w_{j}=y}1-Y_{ij} \\
\gamma'_{\ell}&=\gamma_{\ell}+\sum_{i:z_{i}=x}d_{i\ell}
\end{align}

The natural parameter for each of these distributions is summarized in table \ref{table:natparams}.
To implement coordinate ascent, each update is simply given by the corresponding
entry's expected value under $q$.

\section{Evaluation}

In this section we fit the CASB to two datasets taken directly from the arXiv,
and compare the results to several baseline clusters.

The first dataset consists of all 7\,819 papers uploaded to the arXiv in the astro-ph.CO 
(cosmology) category between 2009 and 2010, and all 621 users who visited the astro-ph.CO 
``/new" or ``/recent" page at least 5 times and read at least 30 articles. We form a 
positive link ($Y_{ij}=1$) between a user and all the papers they read, and a negative
link ($Y_{ij}=0$) between a user and all the papers that appeared on the astro-ph.CO
"/new" or "/recent" pages on the days they visited that they did not read. For documents that appeared on days
when the user did not visit we set $Y_{ij}=\triangle$, making for a total of 1\,090\,588
non-$\triangle$ links and 65\,814 positive links.

We constructed the second dataset analogously. We selected all 6\,677
papers uploaded to hep-th (theoretical high-energy physics) between 2009 and 2010,
and all 1\,449 users satisfying the criteria above, who additionally read at least 70 articles. We defined $Y$ in the same manner as above, leading to 4\,579\,019
non-$\triangle$ links and 318\,703 positive links.

We considered two standard methods of representing documents to illustrate the flexibility
of the model.  One representation was simply bag-of-words (BOW), treating the document as a vector
counting each of the words in its abstract, so $d_{i\ell}$ represents how many times the $\ell$th word
appeared in the abstract of the $i$th document, giving rise to document vector $d_i$. We 
truncated the abstract vocabulary to remove highly uncommon words, leading to a vocabulary of size
4\,951 for astro-ph.CO documents and 3\,282 for hep-th documents.
(We also did evaluations with the full text of the papers, but were forced to more significantly truncate the size of the vocabulary to improve speed.  We found better performance using abstracts than with full texts, using a truncated vocabulary).

For the other representation we preprocessed the data by fitting latent Dirichlet allocation,
and treating documents as counts over topics. That is, for the $i$th document with content vector
$d_i$ we let $d_{i\ell}$ represent the number of times a word from the $\ell$th topic appeared in
the abstract. In our experiments we fit latent Dirichlet allocation with 50 topics.

We chose to focus on the astro-ph.CO and hep-th categories because physicists who study cosmology
and high-energy physics are some of the heaviest users of the arXiv.  Many of these users visit the website daily
to stay up-to-date with the research community.  As such, these categories have vast user data
that is highly representative of the many subcommunities.  These frequent users are also the primary target
of the recommender system currently being built.

To validate the quality of the CASB clusters, we introduced several benchmarks.

In \cite{NallapatiMMSBLDA2008} Nallapati et. al introduce the Link-PLSA-LDA (L-P-LDA) model.
L-P-LDA is a graphical model that combines latent Dirichlet allocation (LDA) \cite{Blei2003LDA}
and the mixed-membership stochastic blockmodel (MMSB) \cite{airoldi2009mixed}. It learns latent
vector representations of the documents satisfying both the topic structure learned by LDA
and the community structure learned by MMSB.  Of all the benchmarks this one is most similar to ours,
in that the latent variables capture both community structure and node content. We applied
this model to both arXiv datasets, treating users as documents with empty content. Once the
vector representations were learned, KMeans was used to arrive at clusters.

In \cite{Gopalan2014Poisson} Gopalan et. al introduce CTPF, a generative model of document
and reader preferences. CTPF learns user preference vectors and document topic vectors in
the same latent space, for the purpose of recommendations.  In this model a rating $r_{ud}$
between a user $u$ and document $d$ is drawn according to 
\begin{equation}
r_{ud} \sim \mathrm{Poisson}(\eta_u^T ( \theta_d + \epsilon_d))
\end{equation}
for user preference vector $\eta_u$, document topic vector $\theta_d$ and 
a small offset vector $\epsilon_d$. The purpose of this model is to learn a latent
space for effective document recommendation, this model is similar to ours in
that it explicitly considers the interactions between users and items, as well
as item contents. We fit this model to the two arXiv datasets and again used KMeans
to go from vector representations to clusters.

We looked to \cite{CombineLinkContent2009} for a benchmark that explicitly learns
clusters, based on link data and document content. Yang et. al propose PCL, a model
of link formation where the probability of a link depends on a node's latent cluster
membership (similar to our setting with the CASB). They extend this model to PCLDC
by discriminatively incorporating content: if $x_i\in\mathbb{R}^d$ is the $i$th
node's content, they assume
\begin{equation}
p(z_i=k) = \frac{\exp(w_k^T x_i)}{\sum_l \exp(w_l^T x_i)}
\label{eq:discriminative}
\end{equation}
where $w_k\in\mathbb{R}^d$ is a weight-vector associated with the $k$th cluster.
Since this model requires that each node be associated with a content-vector,
we trained this model on the arXiv data by assigning each user's content to
be the $0$-vector.  Unfortunately this seriously hindered the performance of the
PCLDC model as a benchmark.  It would be possible to modify this model to
handle nodes without content, but we did not do so.

As another benchmark, we trained 50 dimensional article vectors with the PV-DBOW
method described by Dai et al. \cite{Dai}.  These article vectors are trained
to be predictive of the text within the article using a hierachical softmax
estimation of the log-linear objective,
\begin{equation}
    P( w | a ) = \frac {\exp\left( v_w \cdot v_a \right)}{Z_a} 
    \quad Z_a = \sum_w \exp \left( v_w \cdot v_a \right) .
\end{equation}
Article vectors trained in this way have proven useful to semantic analysis
tasks \cite{parvec} as well as retaining semantic similarity of the articles
\cite{Dai}. The article vectors were trained on the text extracted from the
pdfs of the articles, after lowercasing the text and inserting word boundaries
at each non-alphanumeric character.  Any `word' appearing less than 30 times
was cut from the vocabulary.  After training, the article vectors were clustered
according to spherical k-means as described in Coates et al. \cite{Coates}.

To improve the utility of the article vectors, in the larger training
example, the astro-ph.CO and hep-th articles were augmented with a set of 98\,392
articles chosen to be representative of all of the categories on the arXiv.

Finally, we also used KMeans on the LDA vectors as a benchmark.

In order to select the proper value of $K$, we fit the CASB to each dataset
for $K=2,\dots,10$, and for each of the learned clusters we calculated the
ELBO as in \ref{eq:elbo}. We selected the largest value of $K$ that contributed
at least a 5\% increase to the ELBO. For both datasets this resulted in $K=6$.

In addition, we present a qualitative evaluation of the model applied to a third dataset.
This dataset consists five years of conference proceeding data from the annual
INFORMS conference \cite{informs}, the largest conference of its kind for practitioners of operations 
research. Papers at INFORMS
are presented in sessions, where a session chair will choose three or four papers relevant
to the session's subject.  We treat the set of authors as the users in our model, such that
each author has a positive link to every paper presented in the same session as the author's
own paper and a negative link to every other paper.

\subsection{Community Discovery on INFORMS}

The INFORMS dataset is rich for study because the conference represented has a large number of sessions and presented papers (more than 1\,000 sessions, and just under 4\,000 papers \cite{informs}). It includes many distinct research subcommunities, which CASB is designed to discover.  It is also a field that is very applied in nature, and one in which the authors have expertise, making evaluation of the clusters easier than for the two physics datasets.
We trained the CASB on this dataset, setting $K=5$ arbitrarily.

To evaluate the quality of these clusters, we formed word cloud visualizations of
the frequently occurring words in each cluster.  More specifically, for each word $w$ we formed
the scalars $w_1,\dots,w_K$ such that $w_i$ is the proportion of papers containing the word
$w$ belonging to the $i$th cluster.  Then, in the word cloud corresponding to cluster $i$, the
weight for the $w$th word is given by $w_i$.  This weighting scheme ignores popular stop words
since their distribution will be uniform across all clusters, whereas words that frequently
occur in the $i$th cluster but do not occur in other clusters will have high weight.  We limit
ourselves to words appearing in more than 50 papers. 

\begin{figure*}
    \begin{subfigure}{.5\textwidth}
        \centering
        \includegraphics[width=.8\linewidth]{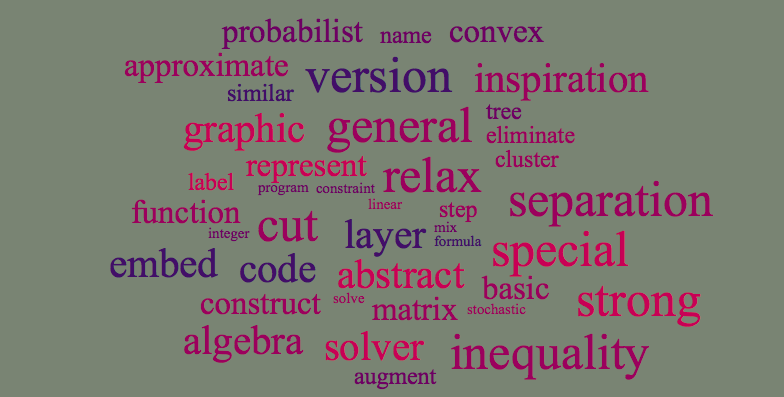}
        \caption{Optimization word cloud}
    \end{subfigure}%
     \begin{subfigure}{.5\textwidth}
        \centering
        \includegraphics[width=.8\linewidth]{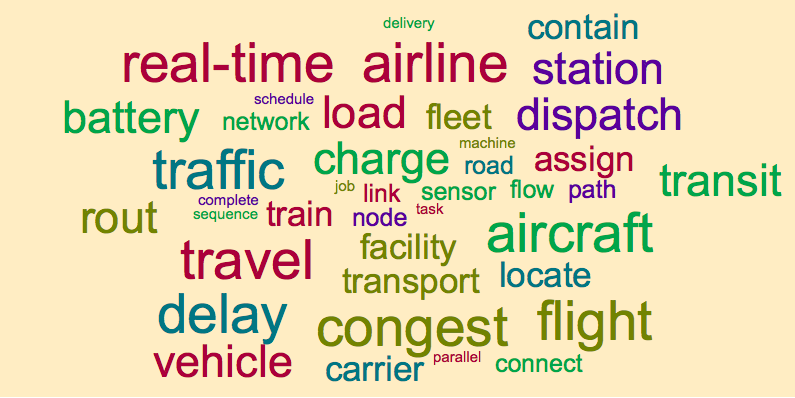}
        \caption{Transporation word cloud}
    \end{subfigure}%
    \caption{Word clouds demonstrating the research communities learned by CASB within the INFORMS dataset.  Each word cloud corresponds to a distinct community, and shows words whose relative frequency are high in that communities' papers.  This qualitative result shows that CASB is able to distinguish meaningful research communities in the INFORMS dataset.}
    \label{fig:fig}
\end{figure*}

Figure \ref{fig:fig} displays two such word clouds. One is clearly representative of the
mathematical optimization research community with words such as ``inequality", ``separation" and
``relax". The other is clearly representative of the transportation logistics community, with
words such as ``airline", ``congest" and ``real-time". We have set up a small website
for a more complete view of the clusters,
as well as the other three word clouds\footnote{peter-i-frazier.github.io/navigate-informs}.

While purely qualitative, these word clouds show that the CASB is able to retrieve real-word
research communities with high accuracy.

\newpage
\subsection{Capturing Misplaced Papers}

This evaluation focuses on two astrophysics subcategories on the arXiv:
Cosmology and Nongalactic Astrophysics (astro-ph.CO);
and Astrophysics of Galaxies (astro-ph.GA).

In creating these categories, the arXiv administrators' intention was for all papers about galactic astrophysics to go to astro-ph.GA.
However, in the past, a significant portion of the astrophysics community had a
different interpretation: astro-ph.GA was for papers discussing our galaxy, the
Milky Way, while papers discussing other galaxies should go to astro-ph.CO.

In late 2013, arXiv.org's moderators began enforcing their interpretation of these two subcategories,
recategorizing papers about galaxies from astro-ph.CO to astro-ph.GA
\cite{ginsparg2014communication}.

We hypothesized that the research communities interested in nongalactic and galactic papers differ, as do the words in their papers, and so the CASB should be able to separate older papers from astro-ph.CO into those nongalactic papers that were correctly submitted to astro-ph.CO, and those galactic papers that should have been submitted to astro-ph.GA. Moreover, it should be able to do this in an unsupervised way, based only on usage and item content, without being given examples of papers in each class.

To test this hypothesis, we fit the CASB and each of the benchmarks to our cosmology dataset consisting of papers submitted to astro-ph.CO over 2009--2010, setting $K=2$. 
We then compared each of these clusterings to a ground truth classification of papers (described below) into those that were properly submitted to astro-ph.CO, and those that should have been submitted to astro-ph.GA.

To create our ground truth, we used a Naive Bayes classifier trained on papers appearing in the arXiv in late 2013 and early 2014, which were manually reclassified by the arXiv moderators.  We then ran this Naive Bayes classifier on the papers in our 2009--2010 dataset.  Note that, although the Naive Bayes classifier is able to automatically classify papers as to whether they belong in astro-ph.GA or astro-ph.CO with high accuracy, this classifier required hand-curated training labels from the arXiv moderators, in the form of correct classifications of a large number of papers from 2013 and 2014.  In contrast, in this evaluation, CASB and the benchmark methods do not have access to this training data, and instead must make a determination based only on what was available in 2009--2010.

\begin{figure*}
    \begin{subfigure}{.25\textwidth}
        \centering
        \includegraphics[width=.9\linewidth]{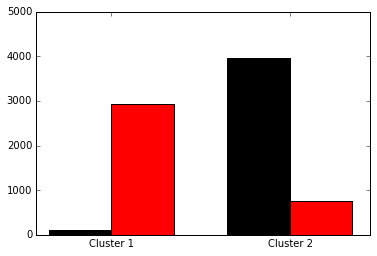}
        \caption{CASB with bag-of-word representations}
    \end{subfigure}%
    \begin{subfigure}{.25\textwidth}
        \centering
        \includegraphics[width=.9\linewidth]{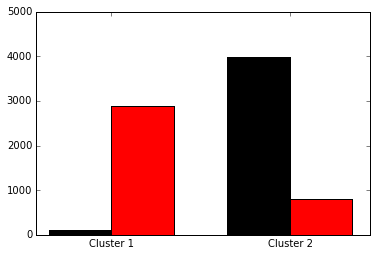}
        \caption{CASB with LDA representations}
    \end{subfigure}%
    \begin{subfigure}{.25\textwidth}
        \centering
        \includegraphics[width=.9\linewidth]{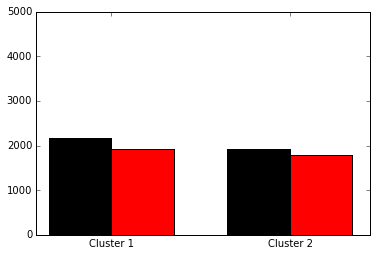}
        \caption{PV-DBOW}
    \end{subfigure}
    \begin{subfigure}{.25\textwidth}
        \centering
        \includegraphics[width=.9\linewidth]{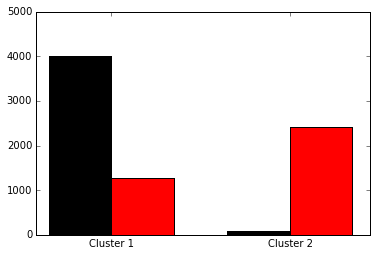}
        \caption{PV-DBOW with auxiliary data}
    \end{subfigure}%
    \\
    \begin{subfigure}{.25\textwidth}
        \centering
        \includegraphics[width=.9\linewidth]{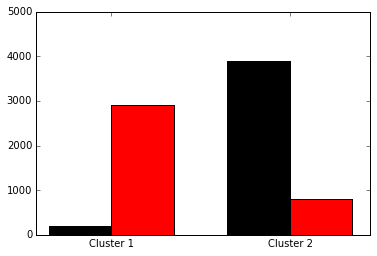}
        \caption{KMeans on Pairwise-link-LDA vectors}
    \end{subfigure}%
    \begin{subfigure}{.25\textwidth}
        \centering
        \includegraphics[width=.9\linewidth]{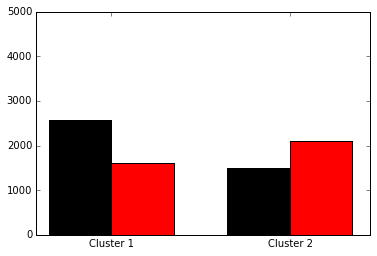}
        \caption{PCLDC}
    \end{subfigure}%
    \begin{subfigure}{.25\textwidth}
        \centering
        \includegraphics[width=.9\linewidth]{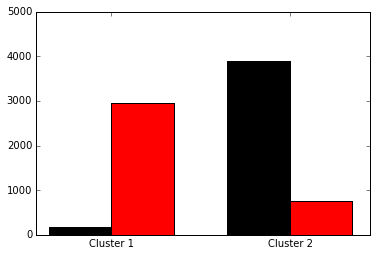}
        \caption{KMeans on LDA vectors}
    \end{subfigure}%
    \begin{subfigure}{.25\textwidth}
        \centering
        \includegraphics[width=.9\linewidth]{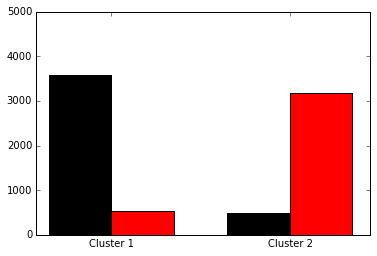}
        \caption{Kmeans on Poisson Factorization vectors}
    \end{subfigure}%
\caption{Distribution of galaxy and cosmology papers amongst clusters of the astro-ph.CO dataset.  The red bars represent cosmology papers and the black bars represent galaxy papers.  A method that performs well puts cosmology papers and galaxy papers in nearly distinct clusters, so that the red bar is much larger than the black bar in one of the clusters, and the black bar is much larger than the red in the other cluster.  }
    \label{fig:astroCOPartition}
\end{figure*}

\begin{table}
\centering
\begin{tabular}{|c|c|} \hline
\textbf{Cluster Type} & \textbf{Misplaced Papers}\\ \hline
KMeans PV-DBOW & 3\,829\\ \hline
PCLDC & 3\,100\\ \hline
KMeans PV-DBOW (auxiliary) & 1\,357\\ \hline
KMeans Poisson Factorization & 1\,024\\ \hline
KMeans Link-PLSA-LDA & 985\\ \hline
KMeans LDA & 936\\ \hline
CASB (LDA docs) & 915\\ \hline
CASB (bag-of-words docs) & 884\\
\hline
\end{tabular}
\caption{Number of misplaced papers for each set of clusters. The
number of misplaced clusters is taken to be the minimum of $g_1 + c_2$
and $c_1 + g_2$ where $g_i$ and $c_i$ are the number of galaxy and cosmology papers in cluster $i$, respectively.}   

\label{table:misplacedClusters}
\end{table}

The distribution of cosmology-classified and galaxy-classified 
papers are presented in Figure~\ref{fig:astroCOPartition}. In 
Table~\ref{table:misplacedClusters} we present the number of misplaced
papers for each clustering scheme. We see that the CASB applied to the bag-of-word representations have the fewest misclustered papers, followed closely by the CASB applied to the LDA representations.

Of note is the fact that the PCLDC clusters have a very high number of misclustered papers, despite considering link presence and document content.  We hypothesize this is because their discriminative incorporation of content does not generalize well to nodes without content.  When the node has zero content, the distribution from (\ref{eq:discriminative}) will be uniform.  Since the links in our datasets only exist between users and documents, PCLDC will have a hard time exploiting the structure of the graph when each user node is uniform across all clusters.

Interestingly, KMeans applied to the LDA representations also has very few misclassified documents. This suggests that there is a lot of signal in the content of the documents pointing to the ground truth communities.  As the table shows, the CASB is able to exploit the user data to discover these communities even more accurately.

\subsection{Author-based Evaluation}

To further evaluate the quality of our clusters we looked to authorship data, as the papers a researcher writes are a strong indicator of the communities to which they belong.

Specifically, for each of our datasets we took the set of authors who had written two or more papers. For each of these authors $a$ we formed the distribution $a_1,\dots,a_K$ where $a_i$ is the proportion of documents $a$ has authored belonging to the $i$th cluster. (This is the same methodology as determining weights for the word clouds).

Now, if a clustering is representative of the underlying community structure, one would expect an author's distribution to be highly concentrated on one or maybe two clusters.  This is because scientific researchers are experts in one or two fields which contain the majority of their work.  They sometimes branch out and write papers in other communities, but this is an infrequent activity.  

To measure this property, we compute the average entropy $\mathcal{H}$ of an author's distribution for each cluster. Entropy is a measure of the disorder of a distribution.  On one extreme, if an author's publications all reside in one cluster the resulting entropy will be $0$.  On the other extreme, if an author's distribution is uniform the resulting entropy will be $\log_2(6) \sim 2.6$ since we fit the model with $6$ clusters.
The entropy for an author $a$ is defined as
\begin{equation}
\mathcal{H}_a = - \sum_i a_i \log_2(a_i).
\end{equation}

\begin{figure*}
    
\centering
    \begin{subfigure}{.4\textwidth}
        \centering
        \includegraphics[width=1.0\linewidth]{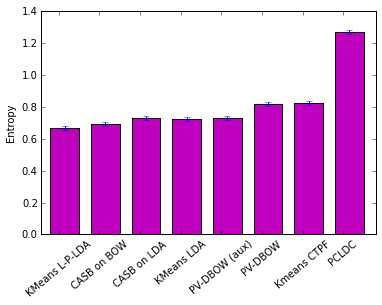}
        \caption{\textbf{astro-ph.CO}}
        \label{sfig:astroph_entropies}
    \end{subfigure}%
    \begin{subfigure}{.4\textwidth}
        \centering
        \includegraphics[width=1.0\linewidth]{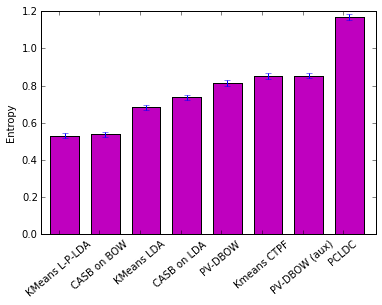}
        \caption{\textbf{hep-th}}
        \label{sfig:hepth_entropies}
    \end{subfigure}%
    \caption{Average author-cluster distribution entropies of the various 
    clusterings for the two arXiv datasets. \textbf{Lower is better.}}
    \label{fig:entropy_barchart}
\end{figure*}

In Figure~\ref{sfig:astroph_entropies} the average author entropy is plotted for each clustering scheme. 
In this plot we see a slight reversal in the quality of the benchmarks compared to the partitioning 
evaluation.  The PV-DBOW clusters trained with auxiliary data has almost identical entropy to CASB 
with LDA representations, and PV-DBOW trained solely on the astro-ph.CO dataset performs better 
than all other benchmarks other than the Link-PLSA-LDA clusters. However, the quality of the CASB 
clusters remain the same: CASB clusters with BOW representations do better than all clusters, 
and CASB clusters with LDA representations are tied for second as previously mentioned.

In Figure~\ref{sfig:hepth_entropies} we see a similar pattern.  The Link-PLSA-LDA clusters have marginally
better performance than the CASB clusters with BOW representations, which are both in turn better
than all other clusters.  The LDA KMeans clusters come in third with slightly better performance
than the CASB clusters with LDA representations.  

It's surprising that the LDA KMeans clusters have better performance than CASB LDA clusters and
KMeans CTPF clusters, since the LDA clusters do not leverage user data at all.  However, we still
see that leveraging user interaction data is worthwhile, as the Link-PLSA-LDA and CASB with
BOW representation clusters have better performance than LDA clusters.

The fact that CASBs learn clusters minimizing the average author-cluster distribution's entropy
again shows that the clusters we are learning are truly representative of the underlying community structure 
in these subcategories.

\subsection{Coreadership Similarities}
\begin{figure*}
    
\centering
    \begin{subfigure}{.4\textwidth}
        \centering
        \includegraphics[width=0.9\linewidth]{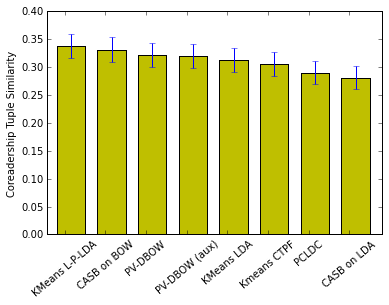}
        \caption{\textbf{astro-ph.CO}}
        \label{sfig:astroph_coreaderships}
    \end{subfigure}%
    \begin{subfigure}{.4\textwidth}
        \centering
        \includegraphics[width=.9\linewidth]{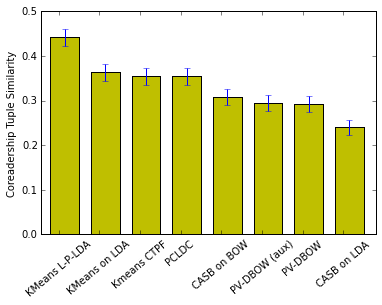}
        \caption{\textbf{hep-th}}
        \label{sfig:hepth_coreaderships}
    \end{subfigure}%
    \caption{Coreadership Similarities of the various clusterings for the two arXiv datasets.}

\label{fig:coreadership_similarity}
\end{figure*}

To understand the CASB clusters better, we wanted to examine the extent to which high coreadership
between two documents determines whether they belong to the same cluster.

Recall for documents $d$ and $b$ with readers $\mathcal{R}_d$ and $\mathcal{R}_b$
respectively, the Jaccard similarity between them is given by 
\begin{equation}
J(d,b) = \frac{|\mathcal{R}_d \cap \mathcal{R}_b|}{|\mathcal{R}_d \cup \mathcal{R}_b|}.
\end{equation}
The Jaccard similarity is a measure of overlap between two sets that is agnostic to their
size.  That is, if one paper has been read by every user, a large intersection between
this paper's readership set with another paper's readership set would not contain
as much signal as a large intersection between two papers with small overall readership.

To evaluate this criteria, we held out 100 users from each of the arXiv datasets, and reran
inference for the CASB and all of the benchmarks. We then selected 3\,000 documents from each of 
the arXiv datasets at random.  For each of these selected documents, we construct another set consisting
of all those documents whose Jaccard similarity with the original is at least $0.5$.  Some of
the originally selected documents did not have Jaccard similarity greater than $0.5$ with any
other documents in the corpus so they were discarded, leaving us with 2\,442 documents from
hep-th and 1\,750 documents from astro-ph.CO. From each of these similarity sets we selected
one document at random, giving rise to a tuple containing the original document and
another document, which together possess Jaccard similarity of at least $0.5$.

After arriving at these tuples with high-coreadership, we simply calculated the proportion
which belonged to the same cluster. The results are summarized in Figure~\ref{fig:coreadership_similarity}.
Of note is that CASB with LDA representations has the lowest coreadership similarity proportion for
both astro-ph.CO and hep-th. Not quite as extreme, the CASB with BOW representations had the second
highest coreadership similarity in the astro-ph.CO dataset, and fifth highest in the hep-th dataset.
These results show that the CASB clusters are not optimizing for high coreadership within clusters.

This can be explained due to the assumptions inherent in the model. Recall the variable $q_{xy}$
represents the probability of a document in cluster $x$ being clicked by a user in cluster $y$.  Hence
if two documents have high Jaccard similarity, it is not necessarily indicative that they arise
from the same community.  Rather, it is possible that both documents belong in separate clusters,
but there is a community of users interested in both of these clusters.  As we see in the author-based
evaluation section, this assumption does not prevent the CASB from learning the true underlying community
structure.

\section{Conclusion}
In this paper we have presented the content-augmented stochastic blockmodel (CASB), 
a probabalistic model of user-item interactions and item content. The cornerstone
assumption of this model is that users and items exist in communities such that
their community memberships determine the probability of an interaction, and the
content of items in the same cluster is generated from the same distribution.
We fit this model to two real-world datasets taken from the arXiv.  We found
that the learned clusters had the highest accuracy in distinguishing between
two real-world communities contained in the dataset, and they gave rise to
author-cluster distributions with low entropy.  Both of these results indicate
that the model's assumptions are valid, and the learned clusters are of high-quality.

\section{Acknowledgments}
JMC, XZ, AAA, YL, and PIF were supported by NSF IIS-1247696.
PIF was also partially supported by NSF CAREER CMMI-1254298, AFOSR FA9550-12-1-0200, AFOSR FA9550-15-1-0038, and the ACSF AVF.  We would like to thank Paul Ginsparg for providing data, and for helpful feedback.  We would also like to thank Thorsten Joachims, Laurent Charlin, and David Blei for helpful feedback.

\bibliographystyle{abbrv}
\bibliography{paper}  
%
%
\end{document}